\documentclass[11pt]{article}

\usepackage[preprint]{acl}
\usepackage{lmodern}

\usepackage{times}
\usepackage{latexsym}

\usepackage[T1]{fontenc}

\usepackage[utf8]{inputenc}

\usepackage{microtype}

\usepackage{inconsolata}
\usepackage{graphicx}

\usepackage{algorithm}
\usepackage{algorithmic}
\usepackage{pifont}
\usepackage{booktabs}
\usepackage{markdown}
\usepackage{mdframed}
\usepackage{url}
\usepackage{makecell}
\usepackage{tcolorbox}
\tcbuselibrary{theorems}

\definecolor{darkblue}{rgb}{0, 0, 0.5}
\hypersetup{colorlinks=true, citecolor=darkblue, linkcolor=darkblue, urlcolor=darkblue}

\newtcolorbox[auto counter, number within=section, list type=subsubsection, list inside=toc]{sectionbox}[2][]{
colback=white!98!gray, colframe=black, 
colbacktitle=white!90!gray, coltitle=black, 
fonttitle=\bfseries,
title={#2}, 
list entry={Comment \thetcbcounter\quad}
}

\newcommand{\cc}{\texttt{\textbf{C-ing Clearly}}}
\newcommand{\noc}{\texttt{\textbf{No C-ing}}}

\newcommand{\multi}{\texttt{\textbf{Multi-turn C-ing Clearly}}}
\newcommand{\llamainstruct}{\texttt{\textbf{Llama-3.1-8B-Instruct}}}

\newcommand{\llamanemotron}{\texttt{\textbf{Llama-3.1-Nemotron-70B-Instruct}}}
\newcommand{\qweninstruct}{\texttt{\textbf{Qwen2.5-14B-Instruct}}}

\title{C-ing Clearly: Enhanced Binary Code Explanations using C code}

\author{
 \textbf{Teodor Poncu\textsuperscript{1,2}},
 \textbf{Ioana Pintilie\textsuperscript{1}},
 \textbf{Marius Dr\u{a}goi\textsuperscript{1}},
 \textbf{Drago\c{s}-Constantin \c{T}\^an\c{t}aru\textsuperscript{1}},
 \textbf{Florin Brad\textsuperscript{1}}
\\
 \textsuperscript{1}Bitdefender,
 \textsuperscript{2}The National University of Science and Technology Politehnica Bucharest
\\
  \texttt{
    \{ipintilie,mdragoi,fbrad\}@bitdefender.com, dan\_teodor.poncu@upb.ro
  }
}
\begin{document}
\maketitle
\begin{abstract}
Large Language Models (LLMs) typically excel at coding tasks involving high-level programming languages, as opposed to lower-level programming languages, such as assembly. We propose a synthetic data generation method named C-ing Clearly, which leverages the corresponding C code to enhance an LLM's understanding of assembly. By fine-tuning on data generated through our method, we demonstrate improved LLM performance for binary code summarization and vulnerability detection. Our approach demonstrates consistent gains across different LLM families and model sizes.
\end{abstract}

\section{Introduction}
LLMs have shown strong performance in high-resource programming languages such as Python or C, while typically under-performing in low-resource languages \citep{orlanski2023measuring,cassano2022multipl} such as assembly. Specializing LLMs in assembly language enhances binary analysis, a key capability for reverse engineering \citep{manuel2024enhancing} and malware detection \citep{alrabaee2022survey}. These specialized models can be obtained by additional pretraining on unsupervised assembly data \citep{Cummins2024meta} or through post-training with \emph{\{assembly code, report\}} supervised data. We focus on the latter by generating better synthetic reports from general-purpose LLMs.

We propose {\cc}, a method for generating synthetic datasets of \emph{\{assembly code, report\}} pairs with LLMs, by leveraging the corresponding C source code. A high-level overview of our method can be seen in Figure~\ref{cing_diagram}. Using both assembly code and C code as input, the {\cc} prompting strategy exploits the generator's knowledge of C code to improve binary code analysis. The C code acts as a high-resource proxy, effectively anchoring the model’s understanding of the assembly.  Our approach is conceptually similar to the work of \citeauthor{Zhang2024BridgeCoderUL} \citeyear{Zhang2024BridgeCoderUL}, which use popular programming languages to improve the code generation abilities for less popular \textit{high-level} programming languages. However, our work focuses \textit{low-level} programming language understanding.
 
Binary Code Summarization (BCS) is the task of summarizing binary programs, usually based on the assembly code recovered from them. Most approaches in BCS target function-level summaries \citep{Ye2023BSC,Jin2023BCS,Lu2024Malsight}, usually by matching source code comments to the corresponding assembly code. However, due to their brevity and implementation-specific nature, source code comments may limit the model’s ability to generate thorough explanations. In contrast, our work focuses on producing more detailed binary code general analysis. 

The general analysis reports created with {\cc} can be further extended with task-specific analyses. To this end, we focus on the vulnerability detection (VD) task. We start from a dataset of \emph{\{C code, vulnerability labels\}} pairs and convert them to \emph{\{assembly code, general analysis + vulnerability analysis\}} pairs. The \emph{vulnerability analysis} locates each \emph{vulnerability label} within the assembly code and describes the associated security risks.

We further experiment with a multi-turn prompting strategy we refer to as {\multi}, in which the LLM analyzes the C code first, then cross-references it with assembly and finally reuses previous completions to obtain the binary analysis.

Using our prompting strategies we generate synthetic data in the form of \emph{\{assembly code, report\}} pairs and fine-tune \llamainstruct. We then evaluate the fine-tuned models' performance on the BCS and VD tasks.

Our contributions can be summarized as follows:

\begin{figure*}[t!]
\includegraphics[width=1.0\linewidth]{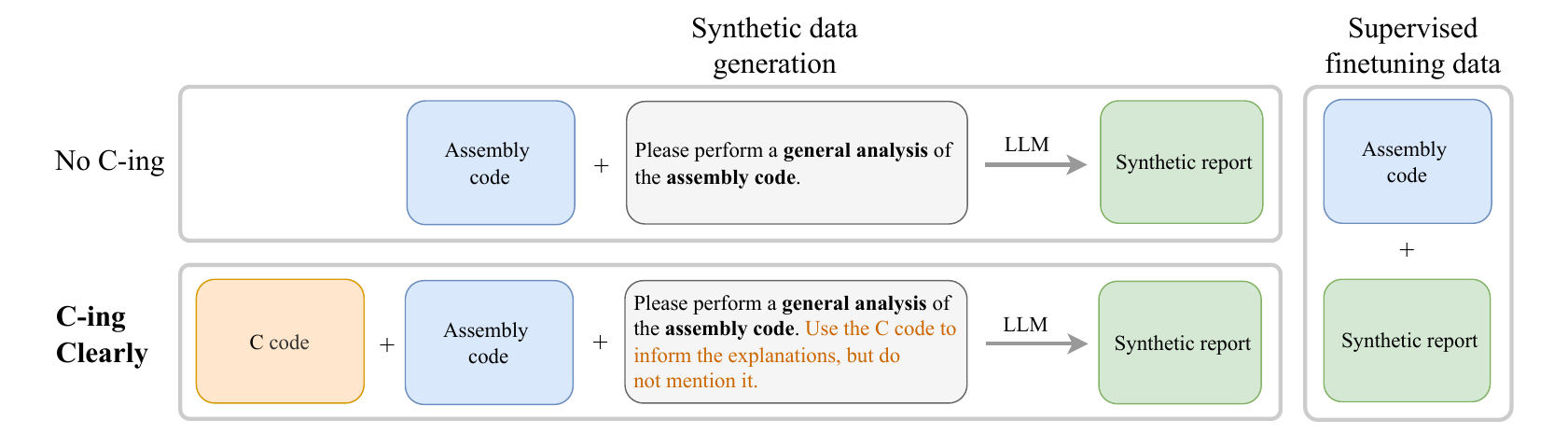}
\caption{{\cc} synthetic data generation versus direct assembly analysis ({\noc}): LLM generates synthetic reports of the assembly code by also using the C code. The \emph{\{assembly, report\}} pairs are then used for supervised fine-tuning.}
\label{cing_diagram}
\end{figure*}

\begin{itemize}
    \item We introduce {\cc}, a synthetic data generation method that addresses the scarcity of in-depth binary code summaries. Our method leverages the LLM's better understanding of C code to obtain more informed assembly annotations.
    \item We show that training with {\cc} improves the LLM's ability to generate assembly code summaries that accurately capture high-level program behavior.
    \item We demonstrate that training with {\cc} also improves vulnerability detection performance, highlighting the potential of our method for other code-related applications. We confirm our findings on different model families as well (\texttt{\textbf{Qwen2.5-14B}}).
\end{itemize}

\section{Method}
\label{method}
In this section we introduce the prompting strategies used in {\cc}, define the tasks along with their dataset construction, and provide model training details.

\subsection{Prompting strategies}

The goal is to generate high-quality completions from a pre-trained LLM $M_{gen}$ for an assembly-related task. Given a compilable C program, we extract its corresponding assembly code and utilize $M_{gen}$ along with a prompt template $P$.

The prompt template $P$ contains two distinct components: the context $c$ and the task-specific instruction $I$. We consider two data-generation setups: {\cc} (the context contains the assembly code \textit{and the original C code}) and {\noc} (the context is assembly code only). We prompt the LLM with $P(context, I)$, which results in a task-specific assembly output. In Figure \ref{cing_diagram}, we illustrate a high-level overview of the two setups. The full set of prompts are listed in Tables~\ref{cing_single}-\ref{no_cing_vuln} in the Supplementary.

We propose an additional strategy we refer to as {\multi}. Instead of directly outputting the assembly report, we engage $M_{gen}$ in a three-turn conversation. In the first turn, the LLM analyzes the C code. The assembly code is cross-referenced with the C code in the second turn. Finally, $M_{gen}$ integrates observations from the previous two turns and outputs an assembly-only analysis. Both setups prevent mentions of the C code in the final output, but the final output of {\multi} may draw on the C code explanations from the intermediate turns.

For both tasks, we use $M_{gen}$=LLama-3.1-Nemotron-70B-Instruct \citep{wang2024helpsteer2preferencecomplementingratingspreferences} to generate the synthetic data. 

\begin{figure*}[ht!]
\includegraphics[width=1.0\linewidth]{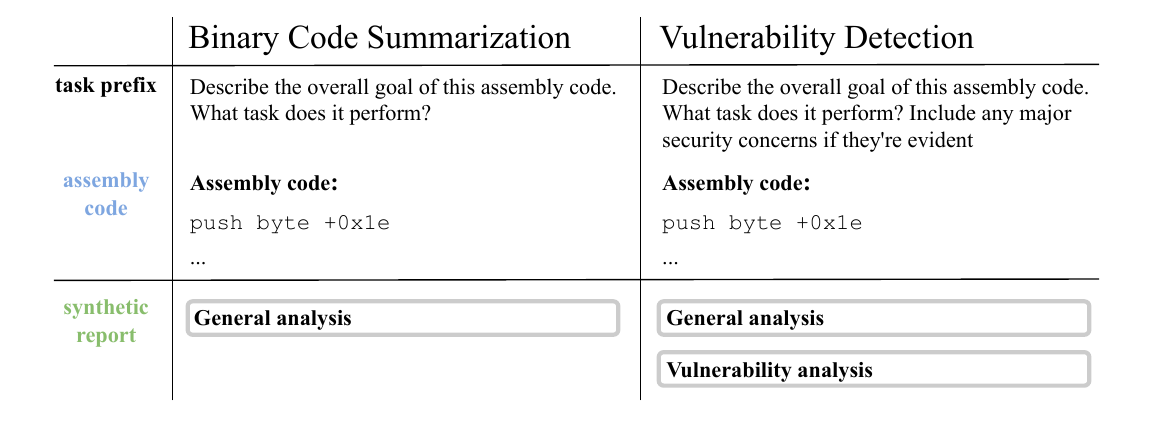}
\caption{Supervised fine-tuning data format for the two tasks}
\label{instruction_tuning}
\end{figure*}

\subsection{Tasks}
\label{tasks}
We use the synthetic data generated with the previously described prompting strategies to perform supervised fine-tuning. We focus on two tasks: Binary Code Summarization (BCS) and Vulnerability Detection (VD). Our goal in BCS is to obtain \emph{general analysis} reports that split the code into assembly blocks, highlight their functionality, and conclude with an overview. For BCS, we fine-tune using \emph{\{task prefix + assembly, general analysis\}} pairs. We hypothesize that retaining the \emph{general analysis} in the output also benefits the VD task.

VD is typically a multi-label vulnerability classification task, performed over popular high-level programming languages such as C++ or Java \citep{Sun2024LLM4VulnAU}. However, we are interested in enhancing the binary code analysis capabilities of an LLM, therefore we cast this as an \textbf{assembly} $\rightarrow$ \textbf{vulnerability labels} generative problem. The vulnerability labels are based on existing taxonomies such as Common Weakness Enumeration (CWE) \footnote{\url{https://cwe.mitre.org}}. We fine-tune the model using \emph{\{task prefix + assembly, general analysis + vulnerability analysis\}} pairs, where the vulnerability analysis is a report containing potential CWE labels, security implications and the relevant assembly code where they occur. In Figure~\ref{instruction_tuning} we depict the structure of the supervised fine-tuning data for both tasks.

The full set of prompts used to generate the training data is provided in both the code repository and the Supplementary for each method: 
\cc\ (Tables~\ref{cing_single}-\ref{cing_single_vul}), \multi\ (Tables~\ref{cing_multi}-\ref{cing_multi_vuln}), and \noc\ (Tables~\ref{no_cing}-\ref{no_cing_vuln}).
For \emph{vulnerability analysis} we use slightly different prompts for vulnerable and non-vulnerable samples. For vulnerable samples we feed the ground truth CWEs and their description, while for non-vulnerable samples we specify that the code is most likely benign. 

\subsection{Supervised fine-tuning datasets}

We generate the supervised fine-tuning data from both vulnerable and non-vulnerable code snippets. Specifically, we start from two datasets, DiverseVul \citep{Chen2023DiverseVulAN} and VDISC \citep{russell2018automated}. They both contain C/C++ functions, collected from open source projects, with vulnerability annotations obtained either through static analysis tools or by crawling security issues and mapping the corresponding commits to affected functions. DiverseVul contains 19K vulnerable functions spanning 150 CWEs. 

After discarding samples with length above 16,384, we obtain \ensuremath{\sim}8.5K vulnerable samples. We keep a held-out test set of 500 samples and use the other samples for training. From the VDISC dataset, we use \ensuremath{\sim}9K non-vulnerable samples for training. For all the C samples, we generate the corresponding assembly code by running the \texttt{gcc -S source.c} command. We generate a supervised fine-tuning dataset for each of the two tasks, as described in subsection~\ref{tasks}. Each dataset contains examples from both DiverseVul and VDISC.

\subsection{Models and training details}
\begin{table*}[t!]
\begin{center}
\begin{tabular}{lrrrrr}
\toprule
\multicolumn{1}{c}{\textbf{vs.}} & \multicolumn{1}{r}{\textbf{Llama}}  &\multicolumn{1}{r}{\textbf{NOC}}  &\multicolumn{1}{r}{\textbf{CC}} &\multicolumn{1}{r}{\textbf{MCC}} &\multicolumn{1}{r}{\textbf{avg. winrate}} \\
\midrule
{\llamainstruct} (\textbf{Llama})              & - & 16.45 & 13.30 & 21.05 & 16.93 \\
{\noc} (\textbf{NOC})            & 82.60 & - & 41.15 & 52.40 & 58.71 \\
{\cc} (\textbf{CC})      & \textbf{85.95} & \textbf{57.09} & - &\textbf{57.45} & \textbf{66.80}\\
{\multi} (\textbf{MCC}) & 78.40 & 46.65 & 40.15 & - & 55.07 \\
\bottomrule
\end{tabular}
\end{center}
\caption{Win rate matrix for all pairwise comparisons between binary code summaries. {\cc} outputs are consistently preferred by the GPT4-o judge.}\label{win-rates}
\end{table*}

\setlength{\tabcolsep}{4pt} 

\begin{table}[t]
\begin{center}
{\small
\begin{tabular}{@{}lrcc@{}}
\toprule
\multicolumn{1}{c}{\shortstack{\textbf{Model}}} &
\multicolumn{1}{r}{\shortstack{\textbf{Output}\\\textbf{tokens}}}  &
\multicolumn{1}{c}{\shortstack{\textbf{Number}\\\textbf{ of}\\\textbf{sections}}}  &
\multicolumn{1}{c}{\shortstack{\textbf{Avg.}\\\textbf{win}\\\textbf{rate}}} \\
\midrule
{\llamainstruct} & 1034.7 & 3.54 & 16.93 \\
{\noc}             & 1074.0 & 3.87 & 58.71 \\
{\cc}        & 873.0 & 3.33 & \textbf{66.80} \\
{\multi} & 894.4 & 2.94 & 55.07 \\
\bottomrule
\end{tabular}
}
\end{center}
\caption{Average length statistics and winrates for binary code summaries generated from different models}\label{length_stats}
\end{table}

We note that training with {\cc} data is model agnostic and benefits any base model that has sufficient knowledge of C, but may lack assembly knowledge. We choose {\llamainstruct} \citep{grattafiori2024llama} as the main base model for our experiments, which we fully fine-tune on each dataset resulted from the {\noc}, {\cc} and {\multi} data generation methods. We fix the training budget to 1B tokens for every experiment.

We also use {\llamainstruct} as the baseline in a zero-shot setting. To generate reports that are similar to our fine-tuned models' outputs, we use the prompt in Table~\ref{no_cing_llama_zero_shot} in the Supplementary. The quality of these reports provides insight into the model’s initial assembly capabilities. 

For model training we set the maximum sequence length to 16,384 and discard examples where the combined length of the prompt and completion exceeds this limit. On average, the assembly code in the supervised datasets is approximately 2,530 tokens long. We set the global batch size to 160. The optimizer is AdamW with a base learning rate of 3e-4. We use linear warmup with cosine annealing, with an initial learning rate of 1e-5 and warmup ratio of 4\%. We perform all the experiments on one NVIDIA 8xA100-80 node.

\section{Binary code summarization}

In this section, we evaluate how well {\llamainstruct} and the fine-tuned models can analyze and explain the assembly code. 

\subsection{Evaluation} 
To evaluate report quality, we employ LLM-as-a-judge \citep{zheng2023judging}. More specifically, we ask GPT-4o \citep{hurst2024gpt} to select the best out of two summaries (each from a different model), based on how accurately it captures the main purpose, high-level functionality and logic of the assembly code. The evaluation prompt is illustrated in Table~13 in the Supplementary. We perform pairwise comparisons between all models.

For each of the 500 test samples, we generate two different outputs from each BCS model, using nucleus sampling \citep{holtzman2019curious} and a temperature of 0.4. This results in a total of 2000 comparisons for each pair of models. The order of the two reports in the evaluation prompt is random.

\subsection{Results}

\newmdenv[
  backgroundcolor=gray!10,
  linecolor=gray,
  topline=false,
  bottomline=false,
  rightline=false,
  leftline=true,
  linewidth=3pt,
  frametitlefont=\bfseries,
  frametitlebackgroundcolor=gray!20,
  frametitleaboveskip=5pt,
  frametitlebelowskip=5pt
]{mdbox}

\begin{figure*}[t]
\begin{mdbox}[frametitle={Judge output: report 1 (No C-ing) vs report 2 (C-ing Clearly)}]
The better report is 2. Reasoning: 

\textbf{Focus on Main Functionality}:

 - Report 2 focuses more directly on the \textit{main functionality} of the assembly code, which is the implementation of a function \texttt{\_Z15tls1\_alert\_codei} that utilizes a jump table based on an input value to return specific codes.
 
 - Report 1, while comprehensive, includes additional details about file and sections, variable and other unrelated functions, which can detract from the main analysis. 

\textbf{Clarity and Conciseness:}

- Report 2 describes the function prologue, conditional branching, jump table usage, and function epilogue \textit{clearly and concisely}. 

- Report 1, although thorough, provides a lot of extra information that might not be directly relevant to understanding the primary logic and purpose of the function.

\textbf{High-Level Insight:}

- Report 2 accurately captures the \textit{high-level insight} that the function maps input values to specific return values using a jump table.

- Report 1 includes additional context and background but doesn't emphasize the \textit{core functionality} as effectively.

Therefore, based on these points, \textbf{Report 2} provides a \textit{more accurate} and focused description of the main purpose, high-level functionality, and logic of the assembly code.

\end{mdbox}
\caption{Excerpts from the output of the GPT4o judge: The model trained with {\cc} data produces a binary code summary that better captures the main functionality of the code.}
\label{fig:nocing_vs_cing}
\end{figure*}

\begin{figure*}[h!]
    \centering
    \includegraphics[width=0.8\linewidth]{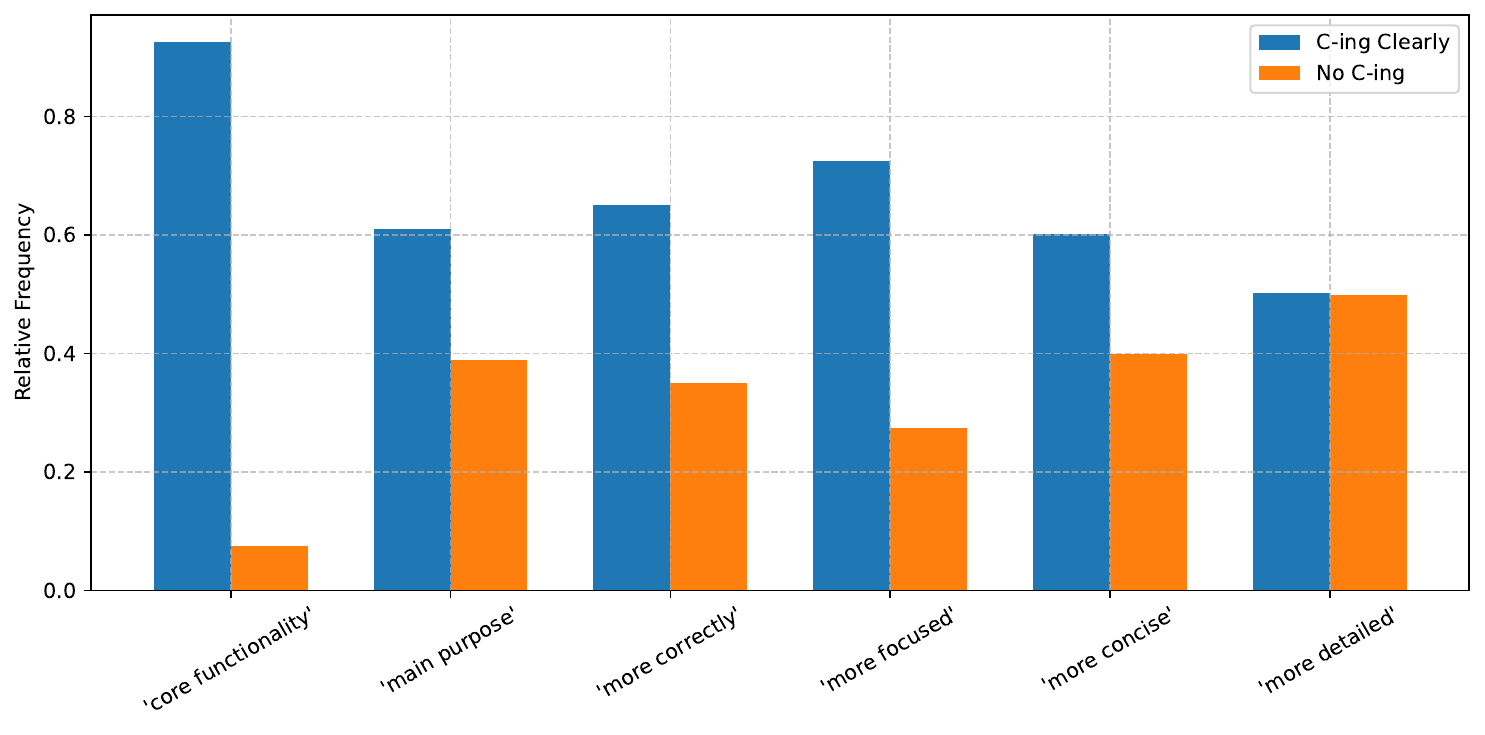}
    \caption{Relative occurrence count for different phrases in the winning report. Mentions of core functionality and correctness are more often linked with {\cc} binary code summaries}
    \label{fig:phrase_winner}
\end{figure*}

The win rate matrix is reported in Table \ref{win-rates}. The model trained with {\cc} data is preferred over all other models, achieving the highest average win rate. Most importantly, {\cc} outputs surpass those from the {\noc} model, showcasing the importance of the C code anchor in the synthetic data generation process.

Outputs from the {\multi} model, although frequently preferred over the baseline outputs, are less competitive. Despite its suboptimal BCS performance, the multi-turn setup has other advantages, discussed in section~\ref{vd_classifier}.

To better understand the summary structure and check for possible LLM judge biases, we analyze in Table~\ref{length_stats} the average response statistics for the different models in relation to their average winrates. {\cc} outputs are both the highest ranked and the shortest on average. In contrast, while {\multi} produces similarly short summaries, the judge ranks them second to last. This suggests that length is not a primary factor in the judge's decisions.

\subsection{Qualitative analysis}

We investigate the answers from the LLM judge to gain insights into the differences between outputs. A sample of the {\cc} output is shown in Table~\ref{cing_clearly_output} in the Supplementary. In Figure~\ref{fig:nocing_vs_cing} we notice that the judge prefers the {\cc} output, highlighting aspects such as core functionality, accuracy and conciseness. The {\noc} output on the other hand is comprehensive but doesn't capture high-level insights as effectively. 

We next examine several key phrases that frequently appear in the judge's responses and are invoked when ranking the outputs. For each phrase, we count how often it occurs when the winning report is produced by either {\cc} or {\noc}. These counts are then normalized by the total number of occurrences of each phrase.

We notice in Figure~\ref{fig:phrase_winner} that phrases linked to high-level functionality ('core functionality', 'main purpose', 'more focused') and correctness ('more correctly') are more often linked to cases where {\cc} is ranked first. Notably, when 'core functionality' is mentioned, {\cc} outputs are ranked first in 92\% of the situations. Phrases linked to brevity are the least discriminative as a winning factor: when 'more detailed' occurs, both models are nearly equally likely to be ranked first. 

\section{Vulnerability detection}\label{VD}
The BCS evaluation indicates improved general understanding of assembly. We also assess whether the datasets resulted from our methods can improve vulnerability detection capabilities of LLMs.  

\subsection{Results}
\label{generative_vd_results}

\begin{figure*}[t!]
\begin{center}
\includegraphics[height=0.16\textheight]{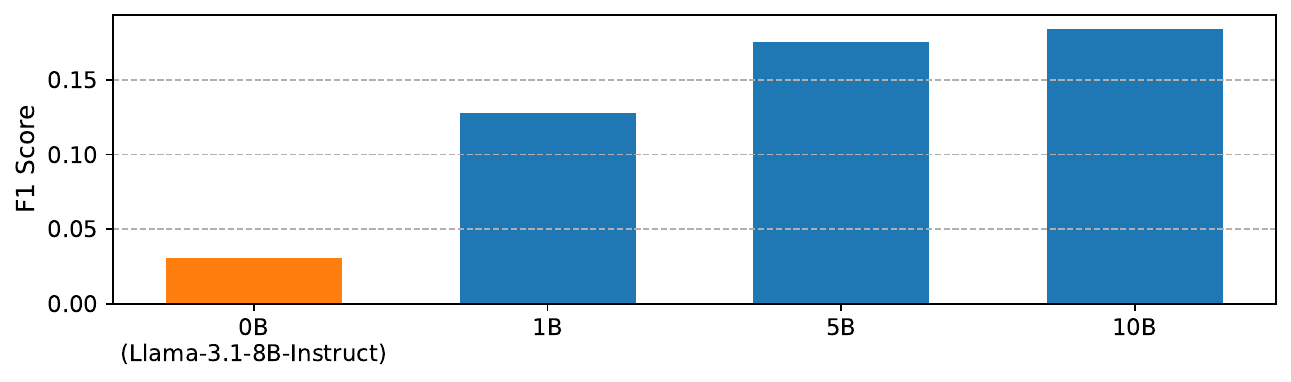}
\end{center}
\caption{F1 score by token budget for {\multi} for vulnerability detection}
\label{f1_by_budget}
\end{figure*}

We use greedy decoding to obtain outputs from models trained for the VD task. CWE label predictions are then extracted from the completions. Finally, we compute micro F1 scores on the test set using ground truth labels from DiverseVul.

\begin{table}[t]
\setlength{\tabcolsep}{5pt} 
\begin{center}
\small{
\begin{tabular}{lrrr}
\toprule
\multicolumn{1}{c}{\bf Model} & \multicolumn{3}{c}{\bf Metrics}  \\
\cmidrule(lr){2-4} 
& $F1$ & $P$ & $R$ \\
\midrule
{\llamainstruct}  & 3.08 & 1.78 & 11.46  \\
{\noc}             & 4.90 & 11.76 & 3.09  \\
{\cc}        & \underline{9.89} & 11.52 & 8.66 \\
{\multi} & \textbf{12.78} & 13.39 & 12.23  \\
\bottomrule
\end{tabular}
}
\end{center}
\caption{F1, Precision and Recall for vulnerability detection for Llama-3.1-8B-Instruct-based models. The best F1 results are shown in bold and second-best results are underlined.}
\label{combined_vd}
\end{table}

We show the results in Table \ref{combined_vd}. While the {\llamainstruct} baseline has a relatively high Recall, suggesting some pre-existing CWE knowledge, it has the worst F1 score. Training with {\noc} shows little improvement over the baseline. 

The model trained with {\multi} data achieves the best F1, Precision and Recall, while the model trained on {\cc} data obtains the second best F1 score. This suggests that our method benefits from multi-turn interactions, such that refining the assembly understanding in two steps leads to better downstream performance. However, for BCS, {\multi} performs worse than {\cc}, suggesting that {\cc} strikes a more effective balance across both tasks.

\subsection{Token budget analysis}
Since fine-tuning on {\multi} data yields the best performing model, we investigate the effect of scaling the fine-tuning token budget in this scenario. The evolution of the performance with a varying token budget is presented in Figure \ref{f1_by_budget}. In addition to the 1B result in Table \ref{combined_vd}, we compute F1 scores after training on 5B and 10B tokens of {\multi} data. Increasing the token budget enhances detection performance, with a notable gain observed when scaling from 1B to 5B tokens.

\begin{table}[h]
\setlength{\tabcolsep}{4pt} 
\begin{center}
{\small
\begin{tabular}{lcrrr}
\toprule
\multicolumn{1}{c}{\bf Model} & \multicolumn{1}{c}{\bf filtered} & \multicolumn{3}{c}{\bf Metrics}  \\
\cmidrule(lr){3-5} & \textbf{train} & $F1$ & $P$ & $R$ \\
\midrule
{\qweninstruct}  & - & 4.0 & 5.2 & 3.2  \\
{\noc}             & \ding{53} & 21.4 & 23.8 & 19.4  \\
{\cc}        &\ding{53} & 18.5 & 25.3 & 14.6  \\
{\noc} & \ding{51} & \underline{21.8} & 22.8 & 20.9 \\
{\cc} & \ding{51} & \textbf{24.0} & 25.7 & 22.6 \\
\bottomrule
\end{tabular}
}
\end{center}
\caption{F1, Precision and Recall for vulnerability detection for Qwen2.5-14B-based models. The best F1 results are shown in bold and second-best results are underlined.}
\label{vd_qwen}
\end{table}

\subsection{Impact of data quality and model}
In this section we address three questions: \textbf{i)} are the results an effect of model distillation? \textbf{ii)} do the results hold for a different model family than LLama? \textbf{iii)} does rejection sampling affect performance?

To answer \textbf{i)} and isolate the effect of the prompt design from the generator's capacity, we evaluate a setting where the base model is the same as the generator model $M_{gen}$. We set $M_{gen}$ = {\qweninstruct}, which addresses question \textbf{ii)} as well.

The ground truth CWEs are provided to the generator model to produce the synthetic VD (assembly, report) data. To answer iii), we filter out reports that do not contain the corresponding CWEs. While completions from {\llamanemotron} have near-perfect match rates between CWEs in the generated analysis and ground truth CWEs (97-99\% match), Qwen completions result in lower matches (58-59\% match rate), requiring rejection sampling. We compare the performance of {\qweninstruct} models when fine-tuned on the original vs the filtered data.

The results are presented in Table~\ref{vd_qwen}. All the finetuned models are better than the zero-shot prompted $M_{gen}$ model, which shows that \textbf{i)} the improvements are not a consequence of distilling from a larger model and \textbf{ii)} the results generalize across different model families.

Regarding \textbf{iii)}, training on filtered synthetic data generally improves downstream performance. {\noc} outperforms {\cc} when training on the raw data, but its performance shows modest improvement with rejection sampling ($+0.4$ F1).
However, we observe that {\cc} benefits more from rejection sampling, showing a larger F1 boost ($+5.5$) and outperforming {\noc}.

\subsection{Multi-label classifier for VD}
\label{vd_classifier}

\begin{table}[t]
\setlength{\tabcolsep}{4pt} 
\begin{center}
{\small
\begin{tabular}{lccc}
\toprule
\multicolumn{1}{c}{\bf Embedding} & \multicolumn{3}{c}{\bf Metrics} \\
\cmidrule(lr){2-4}
& $F1$ & $P$ & $R$  \\
\midrule
{\texttt{\textbf{Base Model}}} & 15.63 {\scriptsize $\pm$ 2.73} & 21.03 {\scriptsize $\pm$ 2.48} & 13.14 {\scriptsize $\pm$ 3.81} \\
{\noc}             & 16.53 {\scriptsize $\pm$ 2.08} & 19.86 {\scriptsize $\pm$ 1.41} & 14.43 {\scriptsize $\pm$ 2.72} \\
{\cc}        & \textbf{17.07} {\scriptsize $\pm$ 1.54} & 17.81 {\scriptsize $\pm$ 2.33} & 17.38 {\scriptsize $\pm$ 4.36} \\
\texttt{\textbf{MCC}} & \textbf{17.07} {\scriptsize $\pm$ 3.04} & 20.84 {\scriptsize $\pm$ 1.20} & 14.93 {\scriptsize $\pm$ 3.30} \\
\bottomrule
\end{tabular}
}
\end{center}
\caption{Results of multi-label classifiers trained over embeddings from different models. MCC stands for \multi, while \textbf{Base Model} refers to Llama-3.1-8B-Instruct.}
\label{vd_discriminative}
\end{table}

Since VD is a classification task, we can train multi-label classifiers on top of assembly-level features extracted from the input program. While not specifically finetuned for assembly representations, we can extract features from the {\llamainstruct}-based models evaluated in Section~\ref{generative_vd_results}.

Specifically, we encode the assembly using the contextual embedding of size 4,096 corresponding to the last token in the \emph{\{task prefix + assembly\}} instruction prompt. The label set spans 135 unique CWEs in the training data. The classifier is a two-hidden-layer MLP, for which we select the best performing detection threshold based on a held-out validation set.
We extract features from both the baseline and fine-tuned models. 

The classification results averaged over 20 independent training runs are presented in Table \ref{vd_discriminative}. As in the generative setup, the best performing models are those trained on {\multi} and {\cc} data, outperforming {\llamainstruct} and {\noc}. 
This suggests that fine-tuning on synthetic data enhanced with C anchors results in more discriminative assembly features for the VD task. While this is not the focus of our work, it indicates a potential direction for future research on optimizing assembly embeddings.

\section{Related Work}
LLMs for assembly are typically designed to learn meaningful assembly representations for tasks such as Binary Code Similarity Detection or to assist with binary analysis tasks such as Binary Code Decompilation or Binary Code Summarization.

\subsection{Models for learning semantic representations of assembly}

To learn assembly representations, both encoder and decoder Transformer-based models \citep{vaswani2017attention} are pretrained then fine-tuned contrastively with different signals. Encoders are trained with Masked Language Modeling (MLM) signal and specialized pretraining tasks such as Jump Target Prediction \citep{Wang2022jtrans}. 

For instance, {\citeauthor{Wang2024clap} \citeyear{Wang2024clap}} introduce CLAP, in which they supervise a RoBERTa encoder \citep{Liu2019roberta} with natural language explanations generated by an LLM from the original source code. ASMA-tune \citep{Wang2025asma} learns a projector module to align the assembly encoder to the decoder's latent space, to benefit from both the encoder's representation learning and the decoder's explanation capability.

{\citeauthor{Jiang2023NovaGL} \citeyear{Jiang2023NovaGL}} add hierarchical self-attention heads to a pre-trained decoder and train it with both next token prediction loss and source code - assembly contrastive loss. This model then transfers well to both embedding tasks and binary analysis tasks such as Binary Decompilation.

\subsection{Models for binary analysis}
LLMs have been more recently used for assembly analysis tasks such as Binary Decompilation and Binary Code Summarization, finetuned from general LLMs \citep{grattafiori2024llama} or code LLMs \citep{Guo2024deepseek}. Efforts to build a foundational LLM for assembly have emerged, with {\citeauthor{Cummins2024meta} \citeyear{Cummins2024meta}} adapting a code LLM \citep{Roziere2023codellama} and further specializing it for downstream tasks such as compiler flag optimization and disassembly into LLVM-IR.

For Binary Decompilation, {\citeauthor{Tan2024llm4decompile} \citeyear{Tan2024llm4decompile}} train end-to-end models called LLM4Decompile to produce the decompiled output based on the assembly input. They outperform classic decompilation tools such as Ghidra \citep{ghidra} and can even further refine Ghidra's output.

Compared to Source Code Summarization, Binary Code Summarization suffers from a scarcity of high-quality training data. To address this, recent papers have introduced large-scale datasets and explored the use of LLMs: BinSum \citep{Jin2023BCS}, CAPYBARA \citep{AlKaswan2023BintT5} and CP-BCS \citep{Ye2023BSC}. These datasets typically provide function-level summaries by pairing source code comments with the corresponding assembly code. In contrast, {\citeauthor{Lu2024Malsight} \citeyear{Lu2024Malsight}} construct MalS, a malware-focused BCS dataset where summaries are obtained with LLMs based on the malicious source code. Our work follows a similar direction and focuses on producing more in-depth binary code summaries via LLMs rather than relying on potentially terse and implementation-specific source code comments.

\subsection{Vulnerability detection}
Prior to LLMs, a common approach to identifying vulnerable code was to use features from a domain-tuned Transformer \citep{thapa2022transformer}. 

Despite leveraging LLMs, recent VD approaches have not yet resulted in consistently high performance \citep{ding2024vulnerability}. Through in-context learning \citep{brown2020language, dong2022survey}, LLMs can be adapted using auxiliary information in the prompt, such as data-flow \citep{zhang2024prompt} or static analysis tools \citep{li2023assisting}.

DLAP \citep{yang2025dlap} employs Retrieval-Augmented-Generation \citep{Liu2020RetrievalAugmentedGF} to integrate the most similar labeled code fragments with static analysis outputs. This retrieved information is further used to select a predefined Chain-of-Thought \citep{wei2022chain, kojima2022large} to guide the vulnerability analysis.

\section{Conclusion}
In this paper, we present a synthetic data generation method named {\cc} to enhance binary code analysis for LLMs. By providing both assembly code and its corresponding C code, we leverage the LLM's in-context abilities to guide it towards better outputs. 

Experiments show that training with {\cc} improves binary code summaries, in terms of correctness, conciseness and understanding of high-level functionality. Moreover, our method improves the performance on the vulnerability detection task, highlighting its effectiveness across related code analysis problems. Finally, ablations on the vulnerability detection task demonstrate that {\cc} is robust across different model families and can help models improve by bootstrapping on self-generated data.

\section{Limitations and Future Work}
Our method leverages high-level C code to support in-context learning of low-level machine representations. We acknowledge that other high-level languages may also aid in reasoning about low-level code such as assembly or bytecode.

Our experiments are based on assembly programs compiled using \texttt{gcc -S}, which applies no optimization. We hypothesize that our proposed method could be also beneficial when applied to optimized assembly code, where the C anchor may help interpret more compact representations.

Additionally, our results are demonstrated on the x86-64 instruction set, without covering other instruction sets such as ARM or RISC-V.

Finally, a practical limitation lies in the size of the assembly programs. These require substantial context capacity from the LLM. We restricted our experiments to inputs with a maximum sequence length of 16K tokens. Future work should explore the scalability and robustness of our approach on longer binaries.

\bibliography{eacl}

\clearpage
\appendix


\section{Supplementary}
\label{sec:appendix}

\begin{table}[h!]
\centering
\begin{minipage}{1.0\textwidth}
\vspace{0mm}    
\centering
\begin{sectionbox}[]{C-ing Clearly: single turn (non-vulnerable)}
    \centering
      \footnotesize
    \begin{tabular}{p{0.97\textwidth} c}
\# Assembly Code Analysis Guide \\
\vspace{0.5em}
\#\# C code \\
```c\\
\textcolor{olive}{\{c\_code\}}\\
```\\
\vspace{0.5em}
\#\# ASM code \\
```asm\\
\textcolor{olive}{\{assembly\_code\}}\\
```\\
\vspace{0.5em}
\#\# Purpose \\
- Provide a \textbf{high-quality, detailed analysis of the given assembly code}, focusing on the \textbf{relationship between C constructs and their compiled assembly} forms.\\
- Importantly, you can use the C code to inform your explanations for the assembly code, but do not mention the C code in your explanation.\\
- Assume that the recipient of your output can only see the assembly code, but if understanding the assembly code is tricky, you can look at the C code to get a better idea of what it's doing.\\
- Ensure the explanation is tailored to experts in assembly analysis.\\

\vspace{0.5em}
\#\# Key Considerations\\
- Account for compiler optimizations and transformations from high-level code to assembly.\\
- Recognize how compile-time optimizations impact the generated assembly code.\\
- Include relevant assembly snippets with detailed inline comments for clarity.\\
- Highlight common patterns and standard practices in assembly.\\

\vspace{0.5em}
\#\# Analysis Structure \\
Start with a general analysis, focusing on detailed and clear explanations for the reader. You will follow it up with a vulnerability-specific analysis, which will probably be empty.\\
\vspace{0.5em}
\textcolor{blue}{\#\# General Analysis}\\
  - Divide the code into logical sections, numbering each (e.g., Section 1).\\
  - Provide a very detailed analysis of each section, deep dive into details\\
  - Do not focus on security issues yet.\\
  - For each section, include 10–15 lines of assembly code with detailed inline comments for key instructions.\\
\#\# General Analysis Conclusion\\
  - Give an overview of what the code is trying to concretely achieve.\\
\vspace{0.5em}
\textcolor{orange}{\#\# Vulnerability-Specific Analysis}\\
It is unlikely the code contains vulnerabilities. You should mention the lack of vulnerabilities in this section. Do not add any additional section headers here.\\
    \end{tabular}
\end{sectionbox}
\vspace{-2mm}
\caption{The prompt for generating the general analysis and the vulnerability analysis in a single turn, for non-vulnerable samples}
\label{cing_single}
\end{minipage}
\end{table}

\begin{table*}[h!]
\centering
\begin{minipage}{1.0\textwidth}
\vspace{0mm}    
\centering
\begin{sectionbox}[]{C-ing Clearly: single turn (vulnerable samples)}
    \centering
      \footnotesize
    \begin{tabular}{p{0.97\textwidth} c}
\# Assembly Code Analysis Guide \\
\vspace{0.5em}
\#\# C code \\
```c\\
\textcolor{olive}{\{c\_code\}}\\
```\\
\vspace{0.5em}
\#\# ASM code \\
```asm\\
\textcolor{olive}{\{assembly\_code\}}\\
```\\
\vspace{0.5em}
\#\# Purpose \\
- Provide a \textbf{high-quality, detailed analysis of the given assembly code}, focusing on the \textbf{relationship between C constructs and their compiled assembly} forms.\\
- Importantly, you can use the C code to inform your explanations for the assembly code, but do not mention the C code in your explanation.\\
- Assume that the recipient of your output can only see the assembly code, but if understanding the assembly code is tricky, you can look at the C code to get a better idea of what it's doing.\\
- Ensure the explanation is tailored to experts in assembly analysis.\\

\vspace{0.5em}
\#\# Key Considerations\\
- Account for compiler optimizations and transformations from high-level code to assembly.\\
- Recognize how compile-time optimizations impact the generated assembly code.\\
- Include relevant assembly snippets with detailed inline comments for clarity.\\
- Highlight common patterns and standard practices in assembly.\\

\vspace{0.5em}
\#\# Analysis Structure \\
Start with a general analysis, focusing on detailed and clear explanations for the reader. You will follow it up with a vulnerability-specific analysis, which will probably be empty.\\
\vspace{0.5em}
\textcolor{blue}{\#\# General Analysis}\\
  - Divide the code into logical sections, numbering each (e.g., Section 1).\\
  - Provide a very detailed analysis of each section, deep dive into details\\
  - Do not focus on security issues yet.\\
  - For each section, include 10–15 lines of assembly code with detailed inline comments for key instructions.\\
\#\# General Analysis Conclusion\\
  - Give an overview of what the code is trying to concretely achieve.\\
\vspace{0.5em}
\textcolor{orange}{\#\# Vulnerability-Specific Analysis}\\
Focus on the following vulnerabilities:\\
\textcolor{olive}{\{CWEs\_and\_their\_description\}}\\
For each unique vulnerability:\\
1. **Description**\\
   - Explain in depth very precisely the vulnerability in assembly.\\
   - Append the relevant assembly chunk near the explanation.\\
   - Reference CWEs as needed and assess security risks.\\
2. **Exploitation**\\
   - Outline potential attack vectors, severity, and provide at least one example exploit scenario.\\
3. **Mitigations**\\
   - Suggest security improvements, including compiler flags or write secure assembly code.\\
    \end{tabular}
\end{sectionbox}
\vspace{-2mm}
\caption{The prompt for generating summary and CWE tags in a single turn for vulnerable samples}
\label{cing_single_vul}
\end{minipage}
\end{table*}

\begin{table*}[h!]
\centering
\begin{minipage}{1.0\textwidth}
\vspace{0mm}    
\centering
\begin{sectionbox}[]{C-ing Clearly: multi-turn (non-vulnerable samples)}
    \centering
      \footnotesize
    \begin{tabular}{p{0.97\textwidth} c}
Analyze the following C code from a functional perspective, focusing on \textbf{high level functionality} and system engineerings aspects:\\
$```$ \\
\textcolor{olive}{\{c\_code\}} \\
``` \\
Please follow these steps in your analysis:\\
Identify the \textbf{main purpose and functionality of the code}. Break down the code into logical sections or functions. \\
For each section: \\
a. Explain its purpose \\
b. Note potential system-level interactions \\
Summarize the overall implications of this code from a systems engineering standpoint. \\

Discuss any assumptions made in the code that could impact security or performance. \\
It's unlikely this code contains any major security issues, but unless there are any obvious issues, mention the lack of vulnerabilities in your explanation.\\
Summarize the overall implications of this code from a security and systems engineering standpoint. \\
Provide a detailed explanation for each step, ensuring a comprehensive understanding of the code's functionality.\\
\hline
Now, \textbf{examine the assembly code produced by the C code you just analyzed}:\\
```\\
\textcolor{olive}{\{asm\_code\}} \\
```\\
Please follow these steps in your analysis:\\
Identify the \textbf{main sections of the assembly code and their correspondence to the C code functions}.\\
For each significant block of assembly:\\
a. Explain its purpose and functionality\\
b. Relate it back to the corresponding C code\\
c. Add line-by-line comments explaining the assembly instructions\\
Highlight any compiler optimizations or unexpected translations from C to assembly.\\
Identify key assembly blocks where specific features or functionalities from the C code are implemented.\\
Discuss the operations regarding memory manipulation.\\
Summarize the key insights gained from examining the assembly version of the code.\\
Provide detailed explanations and comments throughout your analysis, ensuring a \textbf{clear understanding of the relationship between the C code and its assembly representation}.\\
\hline
TURN3
\# Assembly Code Security Analysis Guide\\
\#\# Purpose\\
- Perform an analysis of assembly code, \textbf{based on prior C code analysis}, \textbf{focusing on the relationship between C constructs and their compiled assembly} forms.\\
\#\# Key Considerations\\
- !! Assume the receiver of your output cannot see the previous messages before the TURN3 text and therefore write complete explanations\\
\#\# Analysis Structure\\
Start with a general analysis, focusing on detailed and clear explanations for the reader. You will follow it up with a vulnerability-specific analysis.\\
\textcolor{blue}{\#\# General Analysis}\\
  - Divide the code into logical sections, numbering each (e.g., Section 1).\\
  - Provide a very detailed analysis of each section, deep dive into details. Use your previous analysis to inspire your text, but do not reference C code.\\
  - Do not focus on security issues yet.\\
  - For each section, include 10–15 lines of assembly code with detailed inline comments for key instructions.\\
\#\# General Analysis Conclusion\\
  - Give an overview of what the code is trying to concretely achieve.\\
\textcolor{orange}{\#\# Vulnerability-Specific Analysis}\\
It is unlikely the code contains vulnerabilities. You should mention the lack of vulnerabilities in this section. Do not add any additional section headers here. \\
    \end{tabular}
\end{sectionbox}
\vspace{-2mm}
\caption{The multi-turn prompt for generating summary and CWE tags for non-vulnerable samples}
\label{cing_multi}
\end{minipage}
\end{table*}

\begin{table*}[h!]
\centering
\begin{minipage}{1.0\textwidth}
\vspace{0mm}    
\centering
\begin{sectionbox}[]{C-ing Clearly: multi-turn (vulnerable samples)}
    \centering
      \footnotesize
    \begin{tabular}{p{0.97\textwidth} c}
Analyze the following C code from a functional perspective, focusing on \textbf{high level functionality} and system engineerings aspects:
$```$\textcolor{olive}{\{c\_code\}}``` \\
Please follow these steps in your analysis:\\
Identify the \textbf{main purpose and functionality of the code}. Break down the code into logical sections or functions. \\
For each section: a. Explain its purpose b. Note potential system-level interactions \\
Summarize the overall implications of this code from a systems engineering standpoint. Discuss any assumptions made in the code that could impact security or performance. \\
The code contains the following CWEs:\\
\textcolor{olive}{\{CWEs\_and\_their\_descriptions\}}\\
Knowing that, mention each vulnerable code chunk, and write a detailed step-by-step analysis of the code and vulnerability. A code chunk should containt at least 3 lines of code, and surround it with markdown syntax.\\
For each CWE, write: a. A short description of why the code has that vulnerability and the CWE name. Preface this the text "Vulnerability description:" b. How the CWE might be exploited and how dangerous it would be in practice. Preface this with the text "Possible ways to exploit:" c. Possible ways to solve this issue. Preface this with the text "Mitigations:". Highlight any potential security concerns or optimizations from a systems perspective.\\
Discuss any assumptions made in the code that could impact security or performance. Summarize the overall implications of this code from a security and systems engineering standpoint.\\
Provide a detailed explanation for each step, ensuring a comprehensive understanding of the code's functionality and potential security implications.\\
\hline
Now, \textbf{examine the assembly code produced by the C code you just analyzed}:\\
```\textcolor{olive}{\{asm\_code\}}```\\
Please follow these steps in your analysis:\\
Identify the \textbf{main sections of the assembly code and their correspondence to the C code functions}.\\
For each significant block of assembly: a. Explain its purpose and functionality b. Relate it back to the corresponding C code c. Add line-by-line comments explaining the assembly instructions\\
Highlight any compiler optimizations or unexpected translations from C to assembly.\\
Identify key assembly blocks where specific features or functionalities from the C code are implemented. Discuss the operations regarding memory manipulation. Summarize the key insights gained from examining the assembly version of the code.\\
Provide detailed explanations and comments throughout your analysis, ensuring a \textbf{clear understanding of the relationship between the C code and its assembly representation}.\\
\hline
TURN3
\# Assembly Code Security Analysis Guide\\
\#\# Purpose\\
- Perform an analysis of assembly code, \textbf{based on prior C code analysis}, \textbf{focusing on the relationship between C constructs and their compiled assembly} forms.\\
\#\# Key Considerations\\
- !! Assume the receiver of your output cannot see the previous messages before the TURN3 text and therefore write complete explanations\\
\#\# Analysis Structure\\
Start with a general analysis, focusing on detailed and clear explanations for the reader. You will follow it up with a vulnerability-specific analysis.\\
\textcolor{blue}{\#\# General Analysis}\\
  - Divide the code into logical sections, numbering each (e.g., Section 1).\\
  - Provide a very detailed analysis of each section, deep dive into details. Use your previous analysis to inspire your text, but do not reference C code.\\
  - Do not focus on security issues yet.\\
  - For each section, include 10–15 lines of assembly code with detailed inline comments for key instructions.\\
\#\# General Analysis Conclusion\\
  - Give an overview of what the code is trying to concretely achieve.\\
\textcolor{orange}{\#\# Vulnerability-Specific Analysis}\\
Focus on the following vulnerabilities:\\
\textcolor{olive}{\{CWEs\_and\_their\_description\}}\\
For each unique vulnerability:\\
1. **Description**\\
   - Explain in depth very precisely the vulnerability in assembly.\\
   - Append the relevant assembly chunk near the explanation.\\
   - Reference CWEs as needed and assess security risks.\\
2. **Exploitation**\\
   - Outline potential attack vectors, severity, and provide at least one example exploit scenario.\\
3. **Mitigations**\\
   - Suggest security improvements, including compiler flags or write secure assembly code.\\
    \end{tabular}
\end{sectionbox}
\vspace{-2mm}
\caption{The multi-turn prompt for vulnerable samples}
\label{cing_multi_vuln}
\end{minipage}
\end{table*}

\begin{table*}[h!]
\centering
\begin{minipage}{1.0\textwidth}
\vspace{0mm}    
\centering
\begin{sectionbox}[]{no C-ing Clearly (non-vulnerable samples)}
    \centering
      \footnotesize
    \begin{tabular}{p{0.97\textwidth} c}
\# Assembly Code Security Analysis Guide\\
\#\# ASM code\\
```asm \\
\textcolor{olive}{\{asm\_code\}}\\
```\\
\#\# Purpose\\
- Perform a security analysis of assembly code, identifying vulnerabilities (CWEs), focusing on the compiled assembly sources.\\
- Write very high quality explanations of each section in the general analysis part. \\
\#\# Key Considerations\\
- Map CWE vulnerabilities to assembly.\\
- Account for compiler optimizations and C-to-assembly transformations.\\
- Recognize compile-time optimizations that may appear as missing checks.\\
- Include relevant assembly snippets with each vulnerability.\\
- Group similar vulnerabilities and avoid redundancy.\\
- Standard optimizations may omit some runtime checks.\\
- Fixed-size allocations may lack bounds checks; differentiate optimizations from true vulnerabilities.\\
\#\# Analysis Structure\\
Start with a general analysis, focusing on detailed and clear explanations for the reader. You will follow it up with a vulnerability-specific analysis.\\
\textcolor{blue}{\#\# General Analysis}\\
  - Divide the code into logical sections, numbering each (e.g., Section 1).\\
  - Provide a very detailed analysis of each section, deep dive into details\\
  - Do not focus on security issues yet.\\
  - For each section, include 10–15 lines of assembly code with detailed inline comments for key instructions.\\
\#\# General Analysis Conclusion\\
  - Give an overview of what the code is trying to concretely achieve.\\
\textcolor{orange}{\#\# Vulnerability-Specific Analysis}\\
It is unlikely the code contains vulnerabilities. You should mention the lack of vulnerabilities in this section. Do not add any additional section headers here. \\
    \end{tabular}
\end{sectionbox}
\vspace{-2mm}
\caption{The prompt for generating summary and CWE tags from assembly code only for non-vulnerable samples}
\label{no_cing}
\end{minipage}
\end{table*}

\begin{table*}[h!]
\centering
\begin{minipage}{1.0\textwidth}
\vspace{0mm}    
\centering
\begin{sectionbox}[]{no C-ing Clearly (vulnerable samples)}
    \centering
      \footnotesize
    \begin{tabular}{p{0.97\textwidth} c}
\# Assembly Code Security Analysis Guide\\
\#\# ASM code\\
```asm \\
\textcolor{olive}{\{asm\_code\}}\\
```\\
\#\# Purpose\\
- Perform a security analysis of assembly code, identifying vulnerabilities (CWEs), focusing on the compiled assembly sources.\\
- Write very high quality explanations of each section in the general analysis part. \\
\#\# Key Considerations\\
- Map CWE vulnerabilities to assembly.\\
- Account for compiler optimizations and C-to-assembly transformations.\\
- Recognize compile-time optimizations that may appear as missing checks.\\
- Include relevant assembly snippets with each vulnerability.\\
- Group similar vulnerabilities and avoid redundancy.\\
- Standard optimizations may omit some runtime checks.\\
- Fixed-size allocations may lack bounds checks; differentiate optimizations from true vulnerabilities.\\
\#\# Analysis Structure\\
Start with a general analysis, focusing on detailed and clear explanations for the reader. You will follow it up with a vulnerability-specific analysis.\\
\textcolor{blue}{\#\# General Analysis}\\
  - Divide the code into logical sections, numbering each (e.g., Section 1).\\
  - Provide a very detailed analysis of each section, deep dive into details\\
  - Do not focus on security issues yet.\\
  - For each section, include 10–15 lines of assembly code with detailed inline comments for key instructions.\\
\#\# General Analysis Conclusion\\
  - Give an overview of what the code is trying to concretely achieve.\\
\textcolor{orange}{\#\# Vulnerability-Specific Analysis}\\
Focus on the following vulnerabilities:\\
\textcolor{olive}{\{CWEs\_and\_their\_description\}}\\
For each unique vulnerability:\\
1. **Description**\\
   - Explain in depth very precisely the vulnerability in assembly.\\
   - Append the relevant assembly chunk near the explanation.\\
   - Reference CWEs as needed and assess security risks.\\
2. **Exploitation**\\
   - Outline potential attack vectors, severity, and provide at least one example exploit scenario.\\
3. **Mitigations**\\
   - Suggest security improvements, including compiler flags or write secure assembly code.\\
    \end{tabular}
\end{sectionbox}
\vspace{-2mm}
\caption{The prompt for generating summary and CWE tags from assembly code only for vulnerable samples}
\label{no_cing_vuln}
\end{minipage}
\end{table*}

\begin{table*}[h!]
\centering
\begin{minipage}{1.0\textwidth}
\vspace{0mm}    
\centering
\begin{sectionbox}[]{Zero-shot {\llamainstruct} prompt}
    \centering
      \footnotesize
    \begin{tabular}{p{0.97\textwidth} c}
\# Assembly Code Security Analysis Guide \\
\#\# ASM code
```asm \\
\textcolor{olive}{\{asm\_code\}}\\
```\\
\#\# Purpose \\
- Perform a security analysis of assembly code, identifying vulnerabilities (CWEs), focusing on the compiled assembly sources. \\
- Write very high quality explanations of each section in the general analysis part. \\
\#\# Key Considerations \\
- Map CWE vulnerabilities to assembly. \\
- Account for compiler optimizations and C-to-assembly transformations. \\
- Recognize compile-time optimizations that may appear as missing checks. \\
- Include relevant assembly snippets with each vulnerability. \\
- Group similar vulnerabilities and avoid redundancy. \\
- Standard optimizations may omit some runtime checks. \\
- Fixed-size allocations may lack bounds checks; differentiate optimizations from true vulnerabilities. \\
\#\# Analysis Structure \\
Start with a general analysis, focusing on detailed and clear explanations for the reader. \\
You will follow it up with a vulnerability-specific analysis. \\
\textcolor{blue}{\#\# General Analysis} \\
- Divide the code into logical sections, numbering each (e.g., Section 1). \\
- Provide a very detailed analysis of each section, deep dive into details \\
- Do not focus on security issues yet. \\
- For each section, include 10-15 lines of assembly code with detailed inline comments for key instructions. \\
\#\# General Analysis Conclusion \\
- Give an overview of what the code is trying to concretely achieve. \\
\textcolor{orange}{\#\# Vulnerability-Specific Analysis} \\
For each unique vulnerability: \\
1. **Description** \\
- Explain in depth very precisely the vulnerability in assembly. \\
- Append the relevant assembly chunk near the explanation. \\
- Reference CWEs as needed and assess security risks. \\
2. **Exploitation** \\
- Outline potential attack vectors, severity, and provide at least one example exploit scenario. \\
3. **Mitigations**
- Suggest security improvements, including compiler flags or write secure assembly code. \\
    \end{tabular}
\end{sectionbox}
\vspace{-2mm}
\caption{Zero-shot prompt for the {\llamainstruct} baseline.}
\label{no_cing_llama_zero_shot}
\end{minipage}
\end{table*}

\begin{table*}[h!]
\centering
\begin{minipage}{1.0\textwidth}
\vspace{0mm}    
\centering
\begin{sectionbox}[]{LLM-as-a-judge: evaluation prompt}
    \centering
      \footnotesize
    \begin{tabular}{p{0.97\textwidth} c}
Here is a piece of assembly code: \\
$```$ \\
\textcolor{olive}{\{asm\_code\}} \\
``` \\
The corresponding C/C++ code is: \\
$```$ \\
\textcolor{olive}{\{c\_code\}} \\
``` \\
I will give you two reports by two differents systems. \\

Your task is to choose which of the reports (1 or 2) more correctly describes the main purpose, high-level functionality and logic of the assembly code. \\
Report 1: \\
\textcolor{olive}{\{report1\}}\\
Report 2: \\
\textcolor{olive}{\{report1\}}\\
The experts had no access to the C code so they do not reference it. However, you may use it to better understand what the assembly program does, and use the insights in your judgement. \\
Please finish your answer with 'The better report is x', where x is either 1 or 2, depending on your final choice, or 'none', if you consider the reports equivalent.
    \end{tabular}
\end{sectionbox}
\vspace{-2mm}
\caption{The prompt for BCS evaluation. The judge is GPT-4o \citep{hurst2024gpt}}
\label{judge_prompt}
\end{minipage}
\end{table*}

\begin{table*}[h!]
\centering
\begin{minipage}{1.0\textwidth}
\vspace{0mm}
\centering

\begin{sectionbox}[]{C-ing Clearly output for Binary Code Summarization}
\footnotesize
\begin{tabular}{p{0.97\textwidth}}

\textbf{General Analysis}

\textbf{Section 1: Function Prologue and Setup}

\begin{verbatim}
_Z15tls1_alert_codei:
.LFB1464:
    .cfi_startproc
    endbr64
    pushq %rbp
    .cfi_def_cfa_offset 16
    .cfi_offset 6, -16
    movq %rsp, %rbp
    .cfi_def_cfa_register 6
    movl %edi, -4(%rbp)
\end{verbatim}

\begin{itemize}
\item \textbf{Function Entry}: The \texttt{\_Z15tls1\_alert\_codei} function begins with a prologue, setting up the stack frame and saving the base pointer (\texttt{rbp}).
\item \textbf{Parameter Storage}: The first parameter (\texttt{\%edi}) is stored at \texttt{-4(\%rbp)}, indicating a 32-bit integer parameter.
\end{itemize}

\textbf{Section 2: Conditional Branching and Jump Table}

\begin{verbatim}
cmpl $115, -4(%rbp)
ja .L2
movl -4(%rbp), %eax
leaq 0(,%rax,4), %rdx
leaq .L4(%rip), %rax
movl (%rdx,%rax), %eax
cltq
leaq .L4(%rip), %rdx
addq %rdx, %rax 
notrack jmp *%rax
\end{verbatim}

\begin{itemize}
\item \textbf{Comparison and Branching}: The code compares the input value (\texttt{-4(\%rbp)}) with \texttt{\$115}. If the value is above \texttt{\$115}, it jumps to label \texttt{.L2}.
\item \textbf{Jump Table Calculation}: For values less than or equal to \texttt{\$115}, the code calculates an index into a jump table (\texttt{L4}) using the input value. The table is located at \texttt{.L4(\%rip)}, and the index is computed as \texttt{0(,\%rax,4)}.
\item \textbf{Indirect Jump}: The calculated index is used to perform an indirect jump to the corresponding label in the table. The \texttt{notrack} instruction suggests this is a no-op or a placeholder, potentially indicating a compiler optimization or a specific instruction set architecture (ISA) requirement.
\end{itemize}

\textbf{Section 3: Jump Table and Function Epilogue}

\begin{verbatim}
.L3:
    movl $0, %eax
    jmp .L34 
...
.L2:
    movl $-1, %eax
.L34:
    popq %rbp
    .cfi_def_cfa 7, 8
ret
\end{verbatim}

\begin{itemize}
\item \textbf{Jump Table}: The table contains various labels, each corresponding to a specific return value. The table is defined in the read-only data section (\texttt{rodata}).
\item \textbf{Function Epilogue}: The function concludes with an epilogue, restoring the base pointer and returning control to the caller.
\end{itemize}

\textbf{General Analysis Conclusion}

The analyzed assembly code appears to implement a function that maps input values to specific return values based on a predefined table. The function takes a single 32-bit integer parameter, compares it against a threshold value, and uses a jump table to determine the return value. The code seems to be generated by a compiler, likely GCC, and includes various optimizations and specific instructions for a particular CPU architecture.

\end{tabular}
\end{sectionbox}

\vspace{-2mm}
\caption{Output from a model trained with {\cc} for Binary Code Summarization}
\label{cing_clearly_output}

\end{minipage}
\end{table*}

\end{document}